\documentclass[conference,10pt,letterpaper]{IEEEtran}
\usepackage{amsmath,amssymb}
\usepackage{booktabs}
\usepackage{graphicx}
\usepackage{microtype}
\usepackage[hidelinks]{hyperref}
\title{When Models Disagree with the Majority: Human-Supported Errors in Dynamic Facial Expression Recognition}

\author{
\IEEEauthorblockN{
Yash Pandey\IEEEauthorrefmark{1},
Mahsa Baktashmotlagh\IEEEauthorrefmark{1},
Marius Portmann\IEEEauthorrefmark{1},
Sara Dolnicar\IEEEauthorrefmark{2}\IEEEauthorrefmark{3},
Siamak Layeghy\IEEEauthorrefmark{1}
}
\IEEEauthorblockA{
\IEEEauthorrefmark{1}School of Electrical Engineering and Computer Science,
The University of Queensland, Brisbane, Australia\\
\IEEEauthorrefmark{2}UQ Business School,
The University of Queensland, Brisbane, Australia\\
\IEEEauthorrefmark{3}Faculty of Tourism Studies Turistica,
University of Primorska, Portoro\v{z}, Slovenia
}
}

\begin{document}
\maketitle

{\renewcommand{\thefootnote}{}
\footnotetext{\scriptsize
Submitted to EmotionSense 2027, a workshop held in
conjunction with IEEE PerCom 2027.
This work has been submitted to the IEEE for possible
publication. Copyright may be transferred without notice,
after which this version may no longer be accessible.}}
\setcounter{footnote}{0}
\addtolength{\skip\footins}{5pt}

\begin{abstract}
Annotators can assign different emotion labels to the same facial expression in a video. Yet dynamic facial expression recognition benchmarks typically score each prediction against a single reference label. A prediction may therefore be marked incorrect even when some annotators chose the same emotion. We examine how often these errors match an annotated alternative and whether the matches can be explained by the emotions each model commonly confuses. We evaluate Qwen3-VL-8B-Instruct and Gemma4-12B-IT using zero-shot inference on 11,697 DFEW test videos and 9,172 MAFW videos. To estimate the expected match rate, we shuffle incorrect predictions among videos with the same reference label and similar annotation characteristics. On DFEW videos where annotators disagreed, 61.69\% of the incorrect zero-shot predictions from Qwen3-VL matched a minority vote, compared with 43.13\% after shuffling. Across three fine-tuned Qwen3-VL methods, the observed rates ranged from 69.45\% to 70.51\%, compared with shuffled baselines of 40.83--43.69\%. For Gemma~4, the observed rate was 64.33\%, compared with 50.78\% after shuffling. On MAFW videos with multiple valid emotion labels, Qwen3-VL and Gemma~4 matched an annotated alternative in 29.73\% and 19.44\% of their errors, respectively. The corresponding shuffled baselines were 14.75\% and 11.64\%. We therefore recommend retaining individual annotations and reporting how often benchmark errors match an annotated alternative alongside standard accuracy.
\end{abstract}

\section{Introduction}

Dynamic facial expression recognition (DFER) aims to identify the emotion a person appears to express in a short video. Benchmarks typically evaluate each prediction against a single reference label \cite{jiang2020dfew, liu2022mafw}. However, emotion labels reflect how annotators interpret facial behaviour, and people can interpret the same expression differently~\cite{barrett2019reconsidered}. When the majority vote determines the reference label, standard accuracy hides this disagreement.

Consider a video that six annotators label \emph{neutral} and four label \emph{sad}. A model that predicts \emph{sad} receives the same score as one that predicts \emph{angry}. Both predictions differ from the majority label, but four annotators selected \emph{sad} and none selected \emph{angry}. We call the first prediction a \emph{human-supported error}. The four votes tell us how those annotators perceived the expression. The person's underlying emotion remains unknown.

Previous studies have shown that a single label can obscure differences in human perception~\cite{aroyo2015crowdtruth,uma2021disagreement,chou2025minority}. However, a model that frequently confuses two emotions may often match a minority vote simply because that confusion is common. A video with several annotated alternatives also offers more opportunities for a match. We therefore ask \emph{whether incorrect predictions match alternative annotations for the same video more often than expected from the model's usual errors}.

\IEEEpubidadjcol

We examine this question on two datasets. Dynamic Facial Expression in the Wild (DFEW) records ten individual emotion votes per video~\cite{jiang2020dfew}, while Multi-modal Affective Database for Facial Expression
Recognition in the Wild (MAFW) provides confidence scores for multiple emotions~\cite{liu2022mafw}. We evaluate Qwen3-VL-8B-Instruct~\cite{bai2025qwen3vl} and Gemma4-12B-IT~\cite{gemmateam2026gemma4} on both datasets using zero-shot inference. On DFEW, we also evaluate three fine-tuned Qwen3-VL methods under the official five-fold protocol.

We make three contributions.
\begin{itemize}
    \item We measure how often incorrect predictions match a minority vote on DFEW or a valid alternative emotion on MAFW. We also test stricter criteria for annotation support.
    \item We compare the observed match rates with rates obtained by shuffling incorrect predictions among videos with the same reference label and similar annotation characteristics. This preserves each model's pattern of class confusions within each group.
    \item We examine how model accuracy varies with annotator agreement on DFEW and annotation confidence on MAFW. On MAFW, we also examine how the gap between the two highest confidence scores relates to the likelihood of an alternative match.
\end{itemize}

For both models on both datasets, incorrect predictions matched annotated alternatives more often than expected under matched shuffling. On DFEW, this pattern persisted even when nine of ten annotators agreed. Among its errors on these videos, Qwen3-VL with supervised fine-tuning predicted the emotion chosen by the remaining annotator in 44.99\% of cases, compared with 26.08\% after shuffling.

\section{Related Work}

Dynamic facial expression recognition (DFER) benchmarks preserve more information about human perception than a single accuracy score conveys. DFEW records ten emotion votes for each clip across seven categories~\cite{jiang2020dfew}. MAFW covers eleven categories and provides confidence scores for multiple emotions in the same clip~\cite{liu2022mafw}. Models such as Former-DFER learn how facial expressions change across frames~\cite{zhao2021formerdfer}. DFER-CLIP combines temporal visual features with descriptions of the facial behaviour associated with each emotion~\cite{zhao2023dferclip}. In single-label evaluation, both approaches receive credit when the predicted emotion matches the reference label.

Previous work has shown that annotator disagreement can reflect different interpretations of the same input~\cite{uma2021disagreement,plank2022variation}. Several studies use this information during training. For speech emotion recognition, Upadhyay et al.~\cite{upadhyay2024rater} group raters who tend to interpret utterances similarly and use these groups to define additional training targets. For DFER, MIDAS mixes pairs of videos and their corresponding soft labels, enabling the model to learn from ambiguous DFEW clips~\cite{kawamura2024midas}. In a recent preprint, Wang et al.~\cite{wang2026disagreement} train directly on DFEW vote counts and predict how strongly annotators will disagree on unseen clips. They use this estimate to help the model decide when to withhold a prediction.

Other studies use multiple annotations to evaluate model outputs. Bryant et al.~\cite{bryant2022perception} collected ratings of expression intensity for 1,000 still images and compared the outputs of four facial expression models with the distributions of human ratings. In speech emotion recognition, Chou et al.~\cite{chou2025minority} showed that requiring consensus can exclude ambiguous utterances from the test set. Their aggregation rule retains those utterances and allows several emotion labels to contribute to evaluation.

Our study examines the particular emotion a model selects when its prediction differs from the benchmark label. Consider two clips that each receive eight votes for \emph{neutral}. The remaining two votes are for \emph{sad} in the first clip and \emph{angry} in the second. The degree of agreement is identical, but a prediction of \emph{sad} has human support only in the first clip. By shuffling incorrect predictions among comparable clips, we test whether a model selects the annotated alternative for that particular clip more often than expected from its usual class confusions.

\section{Data and Methods}

\subsection{Datasets and Evaluation Sets}

\textbf{DFEW.} DFEW contains 16,372 annotated video entries. Each entry has ten votes across seven emotion categories (\emph{happy}, \emph{sad}, \emph{neutral}, \emph{angry}, \emph{surprise}, \emph{disgust} and \emph{fear})~\cite{jiang2020dfew}. In 12,059 entries, at least six annotators selected the same emotion, giving it a strict majority. We evaluate the 11,697 videos in the five official test folds. All ten annotators agreed on 3,114 of these videos. The remaining 8,583 received at least one minority vote.

For each fine-tuned method, we train on four folds and test on the remaining fold. Each video therefore appears in the test set exactly once. We use the same videos for zero-shot evaluation without training on DFEW.

\textbf{MAFW.} MAFW contains 10,045 annotated clips~\cite{liu2022mafw}. We evaluate 9,172 videos from its single-expression classification task, which assigns each video one benchmark label from eleven emotion categories. MAFW also provides confidence scores between 0 and 1 for individual emotions. For our analysis, we use MAFW's confidence threshold of 0.50 \cite{liu2022mafw} and treat emotions with recorded scores of at least 0.50 as valid. Of the evaluated videos, 4,295 have at least two emotions meeting this threshold, forming our analysis subset. These scores represent aggregated annotation confidence, whereas DFEW records individual votes.

\subsection{Models and Prediction Methods}

We evaluate two vision--language models, Qwen3-VL-8B-Instruct (Qwen3-VL)~\cite{bai2025qwen3vl} and Gemma4-12B-IT (Gemma~4)~\cite{gemmateam2026gemma4}, using zero-shot inference on both datasets. On DFEW, we also evaluate three fine-tuned Qwen3-VL methods. The first uses supervised fine-tuning (SFT) for emotion classification. The other two share an adapter trained for classification and reasoning using an adaptation of \emph{Multi-Task Cross-Training} (MTCT)~\cite{xu2026rdmldg}. MTCT classification (MTCT-cls) predicts the emotion directly, while MTCT reasoning (MTCT-rsn) uses a reasoning prompt to obtain the prediction.

All methods use only visual input, without audio or captions. We sample sixteen equally spaced frames from each video, or use all frames when fewer than sixteen are available. Training and prompting both vary across the Qwen3-VL methods, which limits what we can attribute to fine-tuning alone. On DFEW, zero-shot Qwen3-VL returned predictions for 11,690 videos, leaving seven missing outputs. SFT and MTCT-cls had 31 and 15 missing outputs, respectively. MTCT-rsn and Gemma~4 returned outputs for all videos. We count missing outputs as errors when calculating standard accuracy.

\subsection{What Counts as a Human-Supported Error?}

\textbf{DFEW.} We define majority agreement as the fraction of the ten annotators who selected the majority emotion. To measure minority support, we consider videos with at least one minority vote. An \emph{eligible error} is a prediction on one of these videos that returns a DFEW emotion other than the majority label. An \emph{Any minority} match occurs when at least one annotator selected the predicted emotion. A \emph{Top minority} match occurs when the prediction matches the minority emotion with the most votes, including ties. As a stricter criterion, we also repeat the analysis requiring at least two annotators to have selected the predicted emotion. For MTCT-rsn, the reported denominator also includes four outputs outside the seven categories, counted as unmatched errors. We check the effect of excluding them.

\textbf{MAFW.} Eligible errors are incorrect predictions on videos with at least two valid emotions. An \emph{Any valid} match occurs when the predicted emotion has a confidence score of at least 0.50. A \emph{Top valid} match occurs when it has the highest score among valid alternatives to the benchmark label. We count the fifteen unparseable Qwen3-VL outputs as unmatched errors and keep them fixed during shuffling. All Gemma~4 outputs were parseable. The different definitions of annotation support limit direct comparisons of match rates across datasets.

\subsection{Stratified Permutation Test}

To estimate how often a model's usual class confusions would produce a match, we randomly shuffle its incorrect predictions among videos with eligible errors. Each video keeps its original human annotations. Shuffling within groups preserves the frequency of each predicted emotion in that group while changing which prediction is paired with each video.

On DFEW, each group shares the same majority emotion, majority vote count and official test fold. As a robustness check, we also match videos on the number of distinct emotions selected by annotators. On MAFW, we group videos by benchmark label and number of valid emotions. In a further check, we also group them by their highest emotion confidence score using intervals of width 0.05.

Consider a DFEW group in which 30\% of a model's incorrect predictions are \emph{sad} and 10\% are \emph{fear}. After shuffling, a video whose only minority label is \emph{sad} has a 30\% probability of an \emph{Any minority} match. For a video whose minority labels are \emph{sad} and \emph{fear}, this probability is 40\%. The expected match rate therefore depends on both the labels the model tends to predict incorrectly and the alternatives annotators selected for each video. We shuffle each method's predictions separately, so two methods can have different baselines on the same dataset. For each shuffle, we divide the total number of matches across all groups by the total number of eligible errors. This gives each eligible error equal weight.

For each test, the mean match rate over 5,000 shuffles gives the \textit{matched-shuffle baseline}. If $b$ shuffles yield a match rate at least as high as the observed rate, the one-sided $p$-value is $(b+1)/5001$ \cite{phipson2010permutation}. The smallest possible value is therefore $1/5001 \approx 0.0002$.

\subsection{Accuracy and Statistical Analysis}

We report \emph{weighted average recall} (WAR), the proportion of videos for which the prediction matches the benchmark label. We also report \emph{unweighted average recall} (UAR), the mean recall across the seven DFEW classes or eleven MAFW classes. We calculate 95\% Wilson confidence intervals (CIs) for observed match rates.

For Qwen3-VL on DFEW, we use logistic regression to examine how majority agreement relates to the odds of a correct prediction, accounting for emotion class and test fold. In the pooled analysis of the three fine-tuned methods, we also adjust for method and cluster standard errors by video because all three methods evaluate the same videos.

For Qwen3-VL on MAFW, we examine how the highest emotion confidence score relates to the odds of a correct prediction, adjusting for benchmark label. Among eligible errors, we examine how the gap between the two highest confidence scores relates to the odds of matching a valid alternative. This analysis adjusts for benchmark label and number of valid emotions. We interpret these relationships as associations.

\section{Results}

\subsection{DFEW Accuracy and Annotator Agreement}

Table~\ref{tab:performance} reports accuracy on all 11,697 DFEW test videos. In the zero-shot setting, Qwen3-VL achieved 52.16\% UAR and 58.13\% WAR, compared with 38.99\% UAR and 46.64\% WAR for Gemma~4. Among the fine-tuned Qwen3-VL methods, MTCT-cls achieved the highest UAR (65.92\%), followed by SFT (65.68\%). Both achieved 75.40\% WAR.

\begin{table}[!htbp]
\centering
\caption{Accuracy on all 11,697 DFEW test videos. Missing predictions count as errors.}
\label{tab:performance}
\footnotesize\setlength{\tabcolsep}{3pt}
\begin{tabular}{lrrr}
\toprule
Model / method & UAR (\%) & WAR (\%) & Missing \\
\midrule
Qwen3-VL zero-shot & 52.16 & 58.13 & 7 \\
Qwen3-VL SFT       & 65.68 & 75.40 & 31 \\
Qwen3-VL MTCT-cls  & 65.92 & 75.40 & 15 \\
Qwen3-VL MTCT-rsn  & 64.81 & 74.45 & 0 \\
\midrule
Gemma~4 zero-shot  & 38.99 & 46.64 & 0 \\
\bottomrule
\end{tabular}
\end{table}

\begin{figure}[!htbp]
\centering
\includegraphics[width=\columnwidth]{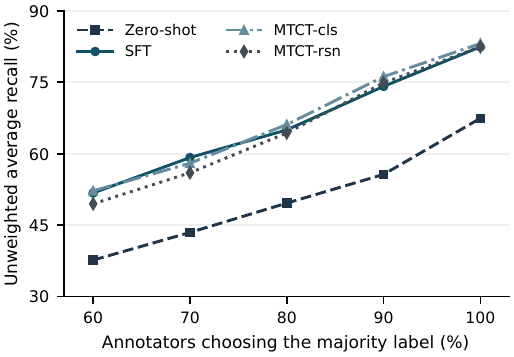}
\caption{Qwen3-VL UAR on DFEW by the percentage of annotators who selected the majority label. Each line represents one prediction method.}
\label{fig:agreement}
\end{figure}

All four Qwen3-VL methods achieved higher UAR on videos with stronger annotator agreement (Fig.~\ref{fig:agreement}). For example, SFT achieved 51.75\% UAR on videos where six of ten annotators agreed and 82.43\% on videos where all ten agreed. Zero-shot Qwen3-VL achieved 37.65\% and 67.43\% on the same two groups.

For the three fine-tuned methods, each ten-percentage-point increase in agreement corresponded to 1.59 times the odds of a correct prediction (95\% CI [1.54, 1.64]), after adjustment for emotion class, test fold and method. The corresponding odds ratio for zero-shot Qwen3-VL was 1.45 (95\% CI [1.40, 1.49]). Excluding \emph{disgust}, the rarest class, gave a fine-tuned odds ratio of 1.59. These comparisons describe how model performance against the majority label varies across groups of videos.

\subsection{Minority-Supported Errors on DFEW}

Table~\ref{tab:dfewerror} considers only eligible errors on videos where annotators disagreed. Zero-shot Qwen3-VL matched a minority vote in 2,562 of 4,153 errors (61.69\%), compared with 43.13\% under matched shuffling. Across the three fine-tuned methods, the observed \emph{Any minority} rates ranged from 69.45\% to 70.51\%, compared with shuffled baselines of 40.83--43.69\%. The 95\% Wilson confidence intervals were 60.20--63.16\% for zero-shot Qwen3-VL and 67.67--71.18\% for SFT.

Gemma~4 matched a minority vote in 3,149 of 4,895 eligible errors (64.33\%, 95\% CI [62.98, 65.66]), compared with a baseline of 50.78\%. Its raw zero-shot match rate was higher than that of Qwen3-VL, but the difference from its own baseline was smaller (13.55 versus 18.56 percentage points). For each of the five methods in Table~\ref{tab:dfewerror}, none of the 5,000 shuffles reached the observed \emph{Any minority} rate ($p=0.0002$).

\begin{table}[!t]
\centering
\caption{DFEW minority-label match rates (\%) among $n$ eligible errors. Shuffle gives the mean rate after shuffling within groups defined by majority emotion, majority vote count and test fold.}
\label{tab:dfewerror}
\footnotesize\setlength{\tabcolsep}{3pt}
\begin{tabular}{lrrrrr}
\toprule
 & & \multicolumn{2}{c}{Any minority}
   & \multicolumn{2}{c}{Top minority} \\
\cmidrule(lr){3-4}\cmidrule(lr){5-6}
Model / method & $n$ & Observed & Shuffle & Observed & Shuffle \\
\midrule
Qwen3-VL zero-shot & 4,153 & 61.69 & 43.13 & 52.81 & 35.45 \\
Qwen3-VL SFT       & 2,642 & 69.45 & 41.48 & 59.01 & 32.45 \\
Qwen3-VL MTCT-cls  & 2,671 & 69.86 & 40.83 & 59.23 & 31.67 \\
Qwen3-VL MTCT-rsn  & 2,787 & 70.51 & 43.69 & 60.60 & 34.77 \\
\midrule
Gemma~4 zero-shot  & 4,895 & 64.33 & 50.78 & 56.26 & 43.20 \\
\bottomrule
\end{tabular}
\end{table}

When we required a match with the most-supported minority emotion (\emph{Top minority}), the rate was 52.81\% for zero-shot Qwen3-VL and 59.01--60.60\% for the fine-tuned methods. Each exceeded its corresponding baseline. Gemma~4 also exceeded its baseline under this criterion (56.26\% versus 43.20\%). Requiring at least two votes for the predicted emotion reduced the Qwen3-VL match rates to 34.07\% for zero-shot and 43.53--44.71\% for the fine-tuned methods.

We next restricted the Qwen3-VL analysis to videos where nine of ten annotators agreed. Each video therefore had exactly one minority vote. SFT matched that vote in 44.99\% of eligible errors, compared with 26.08\% under shuffling. The two MTCT methods had match rates of 46.43--48.08\%, compared with baselines of 24.90--27.14\%. Zero-shot Qwen3-VL matched the minority vote in 44.94\% of errors, compared with 31.01\% under shuffling. The observed rates therefore exceeded their baselines even when each video offered only one alternative.

When we also matched videos on the number of distinct emotions selected by annotators, the shuffled \emph{Any minority} rates rose to 44.67\% for zero-shot Qwen3-VL, 43.67\% for SFT, 43.13\% for MTCT-cls and 45.74\% for MTCT-rsn. All four remained below the observed rates ($p=0.0002$). For Gemma~4, the revised baseline was 51.59\%, compared with the observed 64.33\%. Excluding the four out-of-class MTCT-rsn outputs changed its observed rate from 70.51\% to 70.61\%.

\subsection{Qwen3-VL Errors Across the Full DFEW Test Set and by Emotion}

\begin{figure}[!t]
\centering
\includegraphics[width=\columnwidth]{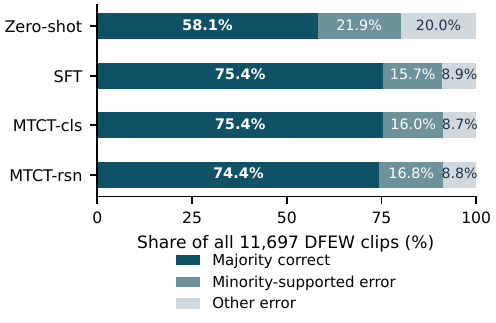}
\caption{Share of all 11,697 DFEW videos for which Qwen3-VL predicted the majority label, predicted a minority label or made another error. Other errors include missing outputs.}
\label{fig:partition}
\end{figure}

Fig.~\ref{fig:partition} uses all 11,697 test videos as the denominator. SFT predicted the majority label on 75.40\% of videos and a minority label on 15.69\%. Other errors, including missing outputs, accounted for 8.92\%. For zero-shot Qwen3-VL, the corresponding shares were 58.13\%, 21.90\% and 19.97\%. Both types of error count as incorrect under standard benchmark scoring. 
Minority-supported errors were less frequent with SFT than with zero-shot inference, but accounted for a larger share of eligible errors.

For SFT, the observed \emph{Any minority} rate exceeded its baseline in all seven majority emotion classes (Table~\ref{tab:classcheck}). The difference was smallest for \emph{disgust} (about 13 percentage points) and largest for \emph{neutral} (33.31 points). The other three Qwen3-VL methods also exceeded their baselines in every class.

\begin{table}[!htbp]
\centering
\caption{Observed and shuffled \emph{Any minority} match rates for Qwen3-VL SFT, grouped by DFEW majority emotion.}
\label{tab:classcheck}
\footnotesize\setlength{\tabcolsep}{4.6pt}
\begin{tabular}{lrrr}
\toprule
Majority emotion & Errors $n$ & Observed (\%) & Shuffle (\%) \\
\midrule
Happy    & 112 & 74.11 & 46.24 \\
Sad      & 321 & 74.77 & 44.62 \\
Neutral  & 780 & 69.10 & 35.79 \\
Angry    & 464 & 59.48 & 35.32 \\
Surprise & 450 & 74.22 & 47.33 \\
Disgust  & 112 & 67.86 & 54.86 \\
Fear     & 403 & 71.22 & 45.46 \\
\bottomrule
\end{tabular}
\end{table}

\subsection{Matches with Alternative Emotion Labels on MAFW}

On MAFW, zero-shot Qwen3-VL achieved 37.90\% UAR and 49.76\% WAR. Gemma~4 achieved 27.77\% UAR and 38.25\% WAR. Table~\ref{tab:mafwerror} reports alternative matches among errors on videos with at least two valid emotions. Qwen3-VL matched a valid alternative in 682 of 2,294 eligible errors (29.73\%, 95\% CI [27.89, 31.63]), compared with a shuffled baseline of 14.75\%. For the highest-scoring valid alternative (\emph{Top valid}), the observed rate was 27.77\% and the baseline was 13.48\% ($p=0.0002$ for both comparisons). Gemma~4 matched a valid alternative in 582 of 2,994 eligible errors (19.44\%, 95\% CI [18.06, 20.90]), compared with a baseline of 11.64\% ($p=0.0002$).

\begin{table}[t]
\centering
\caption{Observed and shuffled \emph{Any valid} match rates among eligible errors on MAFW.}
\label{tab:mafwerror}
\footnotesize\setlength{\tabcolsep}{4pt}
\begin{tabular}{lrrr}
\toprule
Model & Errors $n$ & Observed (\%) & Shuffle (\%) \\
\midrule
Qwen3-VL & 2,294 & 29.73 & 14.75 \\
Gemma~4  & 2,994 & 19.44 & 11.64 \\
\bottomrule
\end{tabular}
\end{table}

When we also matched videos on their highest annotation confidence score, the shuffled \emph{Any valid} rates rose to 15.53\% for Qwen3-VL and 12.17\% for Gemma~4. Both remained below their observed rates. 

\begin{figure}[!htbp]
\centering
\includegraphics[width=\columnwidth]{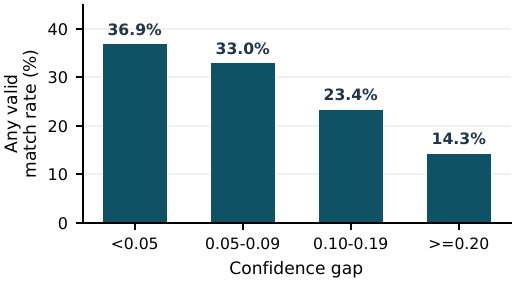}
\caption{Qwen3-VL \emph{Any valid} match rates by the gap between the two highest MAFW annotation confidence scores. The four groups ($<0.05$, 0.05--0.09, 0.10--0.19 and $\geq0.20$) contain 758, 657, 649 and 230 eligible errors, respectively.}
\label{fig:mafw}
\end{figure}

Qwen3-VL errors were more likely to match a valid alternative when the two highest annotation confidence scores were close (Fig.~\ref{fig:mafw}). The match rate was 36.94\% for a confidence gap below 0.05 and 14.35\% for a gap of at least 0.20. After adjustment for benchmark label and number of valid emotions, each 0.10 increase in the gap was associated with 39\% lower odds of a match (odds ratio 0.611, 95\% CI [0.514, 0.725]). In the analysis of benchmark accuracy, each 0.10 increase in the highest confidence score was associated with 1.91 times the odds of a correct prediction after adjustment for benchmark label (95\% CI [1.74, 2.09]).

For Qwen3-VL, the observed \emph{Any valid} rate exceeded its baseline in all eleven MAFW classes (Table~\ref{tab:mafwclasses}). These are comparisons within one dataset. The number of eligible errors ranged from 18 for \emph{happiness} to 538 for \emph{anger}, so the precision of the estimates varies by class.

\begin{table}[!htbp]
\centering
\caption{Observed and shuffled \emph{Any valid} match rates for Qwen3-VL, grouped by MAFW benchmark label.}
\label{tab:mafwclasses}
\footnotesize\setlength{\tabcolsep}{3.5pt}
\begin{tabular}{lrrr}
\toprule
Benchmark label & Errors $n$ & Observed (\%) & Shuffle (\%) \\
\midrule
Anger          & 538 & 7.99  & 3.02 \\
Disgust        & 324 & 16.98 & 8.19 \\
Fear           & 203 & 64.04 & 35.82 \\
Happiness      & 18  & 44.44 & 17.84 \\
Neutral        & 48  & 52.08 & 17.56 \\
Sadness        & 156 & 35.26 & 18.09 \\
Surprise       & 246 & 61.38 & 26.87 \\
Contempt       & 111 & 18.92 & 8.42 \\
Anxiety        & 417 & 33.09 & 16.23 \\
Helplessness   & 113 & 22.12 & 12.71 \\
Disappointment & 120 & 25.83 & 20.99 \\
\bottomrule
\end{tabular}
\end{table}

DFEW records individual votes across seven classes, whereas MAFW records confidence scores across eleven. These differences limit direct comparison of the raw match rates. On both datasets, however, both models selected annotated alternatives more often than expected under their respective matched shuffles.

\section{Discussion and Limitations}

\textbf{Interpreting benchmark errors.}
For both model families on both datasets, incorrect predictions matched an annotated alternative more often than expected under matched shuffling. On DFEW, for example, 69.45\% of eligible SFT errors matched a minority vote. Across the full test set, these errors accounted for 15.69\% of all videos. The first percentage describes the errors SFT made when annotators disagreed. The second shows how often this outcome occurred across the benchmark.

Some emotions are commonly confused, and a video with several annotated alternatives offers more opportunities for a match. The matched shuffles preserve each model's pattern of class confusions within groups of comparable videos. A further check on DFEW considered only videos where nine of ten annotators agreed. Even with a single minority vote, SFT matched that vote in 44.99\% of eligible errors, compared with 26.08\% under shuffling.

\textbf{Performance in context.}
The fine-tuned Qwen3-VL methods achieved 64.81--65.92\% UAR and 74.45--75.40\% WAR on DFEW. Published visual-only results include 63.41\% UAR and 74.43\% WAR for MAE-DFER~\cite{sun2023maedfer} and 66.80\% UAR and 76.68\% WAR for S4D~\cite{chen2026s4d}. Differences in training and evaluation limit direct comparisons. Determining whether these methods also favour annotated alternatives would require their predictions for individual videos.

\textbf{Implications for evaluation.}
We recommend reporting official UAR and WAR alongside two measures when multiple annotations are available. First, report accuracy by annotator agreement or confidence. Second, report the proportion of eligible errors that match an annotated alternative, together with its matched-shuffle baseline and the number of eligible errors. This preserves the benchmark's definition of correctness while showing which errors have human support.

\textbf{Implications for human-centred systems.}
Variation in human annotations highlights the importance of communicating uncertainty in emotion-aware assistants, robots and monitoring systems. Rather than presenting a single emotion label as definitive, such systems could acknowledge alternative interpretations. However, agreement with human annotations does not justify presenting inferred feelings as fact \cite{barrett2019reconsidered}. We evaluated benchmark predictions only. Deployment would require consent, privacy protection and an assessment of whether emotion inference is appropriate for the task.

\textbf{Limits of interpretation.}
Agreement on a label establishes only that the model and an annotator selected the same emotion. An individual annotation may be mistaken, several annotators may share a bias, and a model may select the same label for different reasons \cite{barrett2019reconsidered, uma2021disagreement}. The person's underlying emotion remains unknown. Lighting, facial visibility, pose and scene context could influence label choices. We did not isolate these factors. The models received visual input only, so the role of audio also remains untested. Cultural background may also influence how people express and interpret emotions~\cite{barrett2019reconsidered}. We did not examine cultural differences among the people shown in the videos or among the annotators. How well these findings generalise across cultures or to everyday expressions remains an open question.

\textbf{Experimental scope.}
The four Qwen3-VL methods share one model family, and the two MTCT methods share a trained adapter. Differences in training and prompting limit attribution to either factor. On MAFW, we evaluate only zero-shot predictions. The minority-match analysis excludes DFEW videos with no minority votes. Missing or invalid outputs count as errors in standard accuracy but cannot match minority labels. Future work should evaluate additional model families, annotation reliability and the influence of visual quality.

\section{Conclusion}
When annotators disagree on a facial expression, standard accuracy assigns the same score to two different errors. A model may predict an emotion selected by some annotators, or it may predict an emotion that none of them selected. We examined whether the first type of error occurs more often than expected from the emotions each model commonly confuses. Across both models and both datasets, incorrect predictions matched an alternative annotation for the same video more often than they did when shuffled among comparable videos.

On DFEW videos where annotators disagreed, 61.69\% of eligible zero-shot Qwen3-VL errors matched a minority vote, compared with a shuffled baseline of 43.13\%. Across the three fine-tuned Qwen3-VL methods, the observed rates were 69.45--70.51\%, compared with baselines of 40.83--43.69\%. Gemma~4 matched a minority vote in 64.33\% of eligible errors, compared with its baseline of 50.78\%. We observed the same pattern on MAFW, where annotation confidence scores identify valid alternatives. Qwen3-VL and Gemma~4 matched a valid alternative in 29.73\% and 19.44\% of eligible errors, respectively, compared with baselines of 14.75\% and 11.64\%. All primary permutation tests yielded $p \approx 0.0002$.

These matches show that some benchmark errors reflect an interpretation recorded for that particular video. Such predictions remain errors under the benchmark, and agreement on a label alone cannot establish what the person felt. We therefore recommend that dataset developers retain and release individual annotations wherever possible. Alongside UAR and WAR, researchers can then report how accuracy varies with annotator agreement and how often eligible errors match an annotated alternative relative to a matched-shuffle baseline. Together, these measures show both how reliably a model predicts the reference label and what it predicts when it disagrees.

\bibliographystyle{IEEEtran}
\bibliography{references}

\end{document}